\documentclass[10pt,twocolumn,letterpaper]{article}

\usepackage{wacv}
\usepackage{times}
\usepackage{epsfig}
\usepackage{graphicx}
\usepackage{amsmath}
\usepackage{amssymb}

\usepackage{bm}
\usepackage{multirow}
\usepackage{multicol}
\usepackage{wrapfig}
\usepackage[font=small,labelfont=bf]{caption}
\usepackage{balance}

\usepackage{enumerate}

\usepackage{array}
\newcolumntype{L}[1]{>{\raggedright\arraybackslash}m{#1}}
\newcolumntype{C}[1]{>{\centering\arraybackslash}p{#1}}

\usepackage{authblk}

\def\wacvPaperID{27} 

\wacvfinalcopy 

\ifwacvfinal
\fi

\ifwacvfinal
\usepackage[breaklinks=true,bookmarks=false]{hyperref}
\else
\usepackage[pagebackref=true,breaklinks=true,colorlinks,bookmarks=false]{hyperref}
\fi

\begin{document}

\title{RODNet: Radar Object Detection using Cross-Modal Supervision}


\author[1]{\vspace{-0.5em}Yizhou Wang}
\author[1]{Zhongyu Jiang}
\author[1]{Xiangyu Gao}
\author[1]{Jenq-Neng Hwang}
\author[1]{\protect\\Guanbin Xing}
\author[1,2]{Hui Liu \vspace{-5pt}}
\affil[1]{University of Washington, Seattle, WA}
\affil[2]{Silkwave Holdings Limited, Hong Kong \vspace{3pt}}
\affil[ ]{\tt\small \{ywang26, zyjiang, xygao, hwang, gxing, huiliu\}@uw.edu\vspace{-0.5em}}


\maketitle

\begin{abstract}
Radar is usually more robust than the camera in severe driving scenarios, e.g., weak/strong lighting and bad weather. However, unlike RGB images captured by a camera, the semantic information from the radar signals is noticeably difficult to extract. In this paper, we propose a deep radar object detection network (RODNet), to effectively detect objects purely from the carefully processed radar frequency data in the format of range-azimuth frequency heatmaps (RAMaps). 
Three different 3D autoencoder based architectures are introduced to predict object confidence distribution from each snippet of the input RAMaps. The final detection results are then calculated using our post-processing method, called location-based non-maximum suppression (L-NMS).
Instead of using burdensome human-labeled ground truth, we train the RODNet using the annotations generated automatically by a novel 3D localization method using a camera-radar fusion (CRF) strategy. 
To train and evaluate our method, we build a new dataset -- CRUW, containing synchronized videos and RAMaps in various driving scenarios. After intensive experiments, our RODNet shows favorable object detection performance without the presence of the camera.
\end{abstract}

\section{Introduction}

\begin{figure}[t]
\centering
   \includegraphics[width=.98\linewidth]{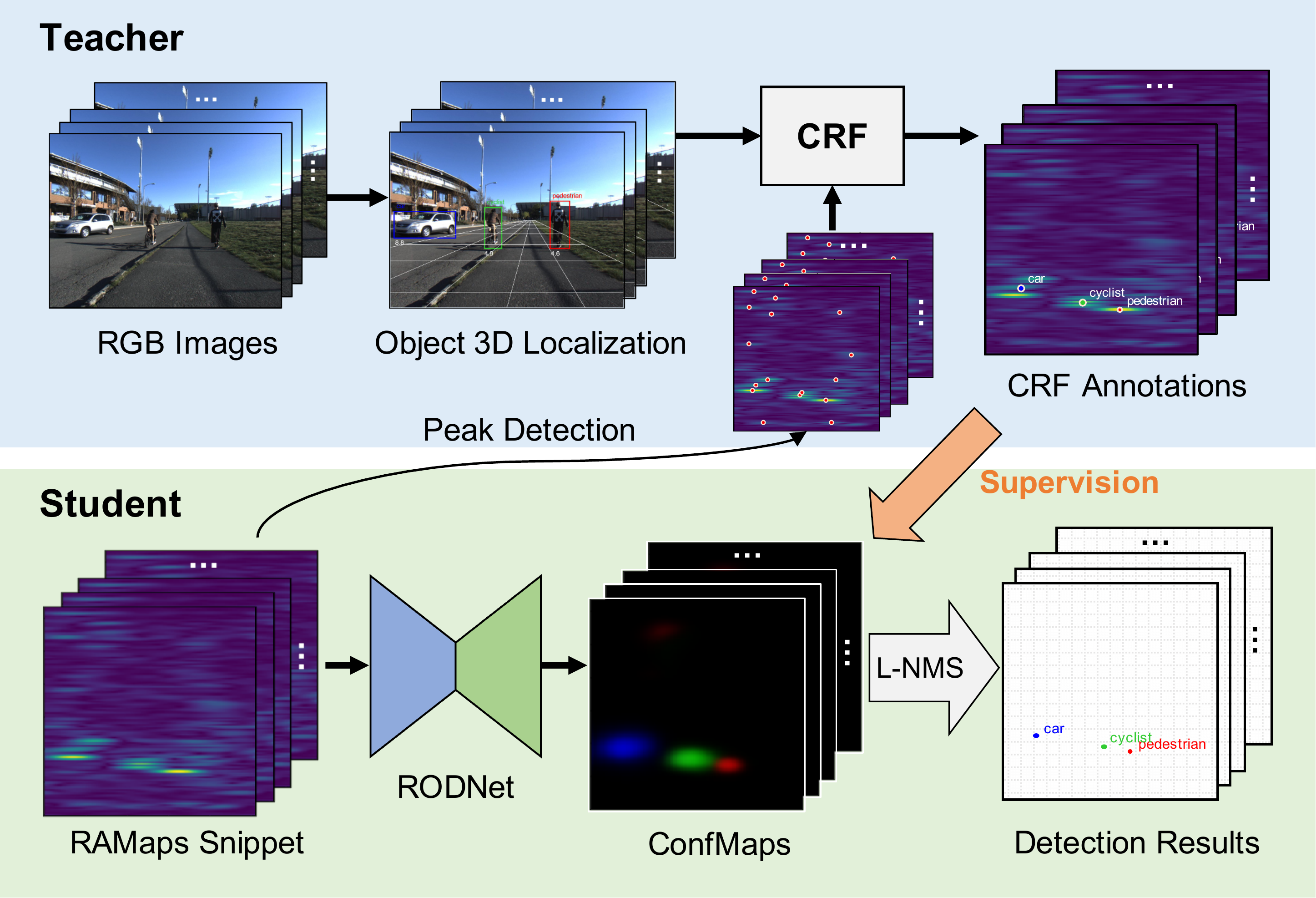}
   \caption{The proposed cross-modal supervision pipeline. Teacher's pipeline first detects and 3D localizes the objects from the RGB images, combined with the detected peaks from the corresponding RAMaps by the proposed camera-radar fusion (CRF) algorithm. Student's pipeline learns to detect objects with radar data (RAMaps) as the input \textbf{only}. }
\label{fig:crosssup_framework}
\end{figure}

In autonomous or assisted driving, a camera can usually give us good semantic understandings of visual scenes. However, it is not a robust sensor under severe driving conditions, such as weak/strong lighting and bad weather, which lead to little/high exposure or blur/occluded images. 
Radar, on the other hand, is relatively more reliable in most harsh environments, e.g., dark, rain, fog, etc. Frequency modulated continuous wave (FMCW) radar, which operates in the millimeter-wave (MMW) band (30-300GHz) that is lower than visible light, thus, has the following properties: 1) great capability to penetrate through fog, smoke, and dust; 2) accurate range detection ability due to the huge bandwidth and high working frequency. 

Typically, there are two kinds of data representations for the FMCW radar, i.e., radar frequency (RF) data and radar points. The RF data are generated from the raw radar signals using a series of fast Fourier transforms (FFTs), and the radar points are then derived from these RF data through peak detection \cite{richards2005fundamentals}. 
Although the radar points can be directly used as the input of the LiDAR point cloud based methods, the radar points are usually much sparser, e.g., less than 5 points on a nearby car, than the point cloud from a LiDAR \cite{nuscenes2019}, so that it is not enough to accomplish the object detection task. Whereas, the RF data can maintain the rich Doppler information and surface texture so as to have the capability of understanding the semantic meaning of a certain object. 
Thus, in this work, we consider the RF data in the range-azimuth coordinates, named RAMaps.

In this paper, we propose a radar object detection method, cross-supervised by a camera-radar fusion algorithm, that can accurately detect objects purely with the radar signal input.
More specifically, we propose a novel radar object detection pipeline, which consists of two parts: teacher and student. The teacher estimates object classes and 3D locations by a reliable probabilistic-driven camera-radar fusion (CRF) strategy to automatically provide annotations for the student. The student takes radar reflection range-azimuth heatmaps (RAMaps) as the input and predicts the object confidence maps (ConfMaps). From the ConfMaps, object classes and locations are inferred using our post-processing method, called location-based non-maximum suppression (L-NMS). The aforementioned pipeline is shown in Figure~\ref{fig:crosssup_framework}.
As for the network architecture of the RODNet, we implement 3D convolutional autoencoder networks based on \cite{zhao2018through} and \cite{newell2016stacked}. Considering different temporal lengths needed for distinguishing different objects, we also propose temporal inception convolution layers, inspired by spatial inception \cite{szegedy2015going}, in our RODNet. 


We train and evaluate the RODNet using our self-collected dataset, called CRUW, which contains about 400K camera-radar synchronized frames with various driving scenarios. 
Our CRUW dataset is, to the best of our knowledge, the first dataset containing synchronized stereo RGB images and RF data for autonomous driving applications. 
To evaluate the performance of our proposed RODNet, without the definition of bounding boxes widely used in image-based object detection on RAMaps, we introduce an evaluation method to evaluate the radar object detection performance in the radar range-azimuth coordinates. With intensive experiments, our RODNet can achieve about $83.76\%$ average precision (AP) and $85.62\%$ average recall (AR) solely based on radar input in various scenarios whether objects are visible or not in cameras.

Overall, our main contributions\footnote{The CRUW dataset and code are available at: \url{https://www.cruwdataset.org/}} are the following:
\begin{itemize}
    \item A novel and accurate radar object detection network, called RODNet, for robust object detection in various driving scenarios, without camera or LiDAR.
    \item A camera-radar fusion (CRF) cross-modal supervision framework for training the RODNet without laborious and inconsistent human labeling. 
    \item A new dataset, named CRUW, is collected, containing synchronized camera and radar data, which is valuable for camera-radar cross-modal research.  
    \item A new evaluation method for radar object detection tasks is proposed and justified for its effectiveness.
\end{itemize}

The rest of this paper is organized as follows. Related works for camera and radar data learning are presented in Section~\ref{sec:relatedworks}. The proposed cross-modal supervision framework is introduced in Section~\ref{sec:cross_modal}, with training and inference of our RODNet being explained in Section~\ref{sec:rod}. In Section~\ref{sec:dataset}, we introduce our self-collected CRUW dataset. Then, the implementation details, evaluation metrics, and experimental results are shown in Section~\ref{sec:experiments}. Finally, we conclude our work in Section~\ref{sec:conclusion}.

\section{Related Works}
\label{sec:relatedworks}

\subsection{Learning of Vision Data}

Image-based object detection \cite{ren2015faster,he2017mask,redmon2018yolov3,lin2017focal} is aimed to detect every object with its class and precise bounding box location from RGB images, which is fundamental and crucial for camera-based autonomous driving. 
Then, most tracking algorithms focus on exploiting the association between the detected objects in consecutive frames, the so-called tracking-by-detection framework \cite{bergmann2019tracking,yang2019video,tang2019moana,wang2019exploit,hsu2019multi,cai2020ia,zhanglifts,hsu2020traffic}. Among them, the TrackletNet Tracker (TNT) \cite{wang2019exploit} is an effective and robust tracker to perform multiple object tracking (MOT) of the detected objects with a static or moving camera. Once the same objects among several consecutive images are associated, the missing and erroneous detections can be recovered or corrected, resulting in better subsequent 3D localization performance. 

Object 3D localization has attracted many interests in autonomous and safety driving community \cite{song2014robust,song2015joint,mousavian20173d,murthy2017reconstructing,ansari2018earth}.
One idea is to localize vehicles by estimating their 3D structures using a CNN, e.g., 3D bounding boxes \cite{mousavian20173d} and 3D keypoints \cite{murthy2017reconstructing,ansari2018earth,kim2017road}. 
Then, they deform a pre-defined 3D vehicle model to fit the 2D projection, resulting in accurate vehicle locations. 
Another idea \cite{song2014robust,song2015joint}, however, tries to develop a real-time monocular structure-from-motion (SfM) system, taking into account the SfM cues and object cues. 
Although these kinds of works achieve favorable performance in object 3D localization, they only work for vehicles since only the vehicle structure information is considered. 
To address this limitation, an accurate and robust object 3D localization system, based on the detected and tracked bounding boxes of objects, is proposed in \cite{wang2019monocular}, claiming that the system works for most common moving objects in the road scenes, such as cars, pedestrians, and cyclists. Thus, we decide to take this 3D localization system as our camera annotation method.

\subsection{Learning of Radar Data}

Significant research in radar object classification has demonstrated its feasibility as a good alternative when cameras fail to provide good results \cite{6042174,8468324,capobianco2017vehicle,kwon2017human,cao2018radar}. With handcrafted feature extraction, Heuel, et al.~\cite{6042174} classify objects using a support vector machine (SVM) to distinguish cars and pedestrians. While, Angelov et al. \cite{8468324} use neural networks to extract features from the short-time Fourier transform (STFT) heatmap. 
However, the above methods only focus on \emph{classification} tasks, that assume only one object has been appropriately identified in the scene and not applicable to the complex driving scenarios.
Recently, a radar object detection method is proposed in \cite{gao2019experiments}, which combines a statistical detection algorithm CFAR \cite{richards2005fundamentals} with a neural network classifier VGG16 \cite{simonyan2014very}. 
But it would easily give many \textit{false positives}, i.e., obstacles detected as objects. Besides, the laborious human annotations required by this method are usually impossible to obtain. 

Recently, the concept of cross-modal learning has been discussed in machine learning community \cite{karpathy2015deep,venugopalan2015sequence,qi2016sketch,jing2019self}. This concept is trying to transfer or fuse the information between two different modalities in order to help train the neural networks. 
Specifically, RF-Pose \cite{zhao2018through} introduces the cross-modal supervision idea into wireless signals to achieve human pose estimation based on WiFi range radio signals. As the human annotations for wireless signals are difficult to obtain, RF-Pose uses a computer vision technique, i.e., OpenPose \cite{cao2017realtime}, to generate annotations for training from the camera. 
However, radar object detection is more challenging: 1) Feature extraction for object detection (especially for classification) is more difficult than human joint detection, which could just classify different joints by their relative locations without considering object surface texture or velocity information; 2) The typical FMCW radars on the vehicles have much less resolution than the sensors used in RF-Pose.
As for autonomous driving, \cite{major2019vehicle} proposes a vehicle detection method using LiDAR information for cross-modal learning. However, our work is different from theirs: 1) they only consider vehicles as the target object class, while we detect pedestrians, cyclists, and cars; 2) the scenario of their dataset, mostly highway without noisy obstacles, is easier for radar object detection, while we are dealing with various traffic scenarios.

\section{Cross-Modal Supervision}
\label{sec:cross_modal}

\subsection{Radar Signal Processing and Properties}
\label{subsec:ramap}


In this work, we use a common range-azimuth heatmap representation, named RAMap, to represent our radar signal reflections. RAMap can be described as a bird's-eye view (BEV) representation, where the $x$-axis shows azimuth (angle) and the $y$-axis shows range (distance).
For an FMCW radar, it transmits continuous chirps and receives the reflected echoes from the obstacles. After the echoes are received and processed, we implement the fast Fourier transform (FFT) on the samples to estimate the range of the reflections. A low-pass filter (LPF) is utilized to remove the high-frequency noise across all chirps. After the LPF, we conduct a second FFT on the samples along different receiver antennas to estimate the azimuth angle of the reflections and obtain the final RAMaps. After transforming into RAMaps, the radar data become a similar format as image sequences, which can be directly processed by an image-based CNN.


Moreover, RF data also has the following special properties to be handled for object detection task. 
\begin{itemize}
\item \textbf{Rich motion information.} 
According to the principle of the radio signal, rich motion information is included. The speed and its law of variation over time consist of surface texture movement details, etc. 
For example, the motion information of a non-rigid body, like a pedestrian, is usually random, while for a rigid body, like a car, it should be more consistent. 
To utilize this motion information, multiple consecutive radar frames need to be considered as the input. 
\item \textbf{Inconsistent resolution.} 
Radar usually has high-resolution in range but low-resolution in azimuth due to the limitation of radar specifications, like the number of antennas, and the distance between them. 
\item \textbf{Different representation.} 
Radar data are usually represented as complex numbers containing frequency and phase information. This kind of data is unusual to be modeled by a typical neural network. 
\end{itemize}

\subsection{Camera-Only (CO) Supervision}
\label{subsec:co_anno}

The annotations for the radar are in the radar range-azimuth coordinates (similar to those of the BEV of a camera). To recover the 3D information from 2D images, we take advantage of a recent work on an effective and robust system for visual object 3D localization based on a monocular camera \cite{wang2019monocular}. 
Even though stereo cameras can also be used for object 3D localization, however, high computational cost and sensitivity to camera setup configurations (e.g., baseline) result in the limitation of the stereo camera localization system.
The proposed system takes a CNN inferred depth map as the input, incorporating adaptive ground plane estimation and multi-object tracking results, to effectively estimate object classes and 3D locations relative to the camera. 

However, the above camera-only system may not be accurate enough after transforming to the radar range-azimuth coordinates because: 1) The systematic bias in the camera-radar sensor system that the peaks in the RF images may not be consistent with the 3D geometric center of the object; 2) Cameras' performance can be easily affected by lighting or weather conditions. Since we do have the radar information available, camera-radar cross calibration and supervision should be used. Therefore, an even more accurate self-annotation method, based on camera-radar fusion, is required for training the RODNet.

\subsection{Camera-Radar Fusion (CRF) Supervision}
\label{subsec:crf_anno}

The camera-only annotation can be improved by radar, which has a plausible capability of distance estimation. The CFAR algorithm \cite{richards2005fundamentals} is commonly used in signal processing community to detect peaks from RAMaps. As shown in Fig.~\ref{fig:crosssup_framework}, a number of peaks are detected from the input RAMaps. However, these peaks cannot be directly used as the supervision because 1) CFAR cannot provide the object classes for the peaks; 2) CFAR usually gives a large number of false positives. Thus, these radar peaks are fused with the above CO supervision using an effective CRF strategy. 

First, the CO annotations are projected from 3D camera coordinates to radar range-azimuth coordinates by 
the sensor system calibration.
After the coordinates between camera and radar are aligned, a probabilistic CRF algorithm is developed. 
The basic idea of this algorithm is to generate two probability maps for camera and radar locations separately, and then fuse them by element-wise product. The probability map for camera locations with object class $cls$ is generated by
\begin{equation*}
\resizebox{0.49\textwidth}{!}{$
    \mathcal{P}^c_{(cls)}(\mathbf{x}) = \max_{i} \left\{ \mathcal{N} \left( \frac{1}{2 \pi \sqrt{|\mathbf{\Sigma}^c_{i(cls)}|}} \exp \left\{ -\frac{1}{2} (\mathbf{x} - \mathbf{\mu}^c_{i})^{\top} (\mathbf{\Sigma}^c_{i(cls)})^{-1} (\mathbf{x} - \mathbf{\mu}^c_{i}) \right\} \right) \right\},
$}
\end{equation*}
\begin{equation}
\resizebox{0.32\textwidth}{!}{$
    \mathbf{\mu}^c_{i} = 
    \begin{bmatrix}
    \rho^c_{i} \\
    \theta^c_{i} 
    \end{bmatrix}, 
    \mathbf{\Sigma}^c_{i(cls)} = 
    \begin{bmatrix}
    \left(d_i s_{(cls)} / c_i \right)^2 & 0 \\
    0 & \delta_{(cls)} 
    \end{bmatrix}.
$}
\end{equation}
Here, $d_i$ is the object depth, $s_{(cls)}$ is the scale constant, $c_i$ is the depth confidence, and $\delta_{(cls)}$ is the typical azimuth error for camera localization. $\mathcal{N}(\cdot)$ represents the normalization operation for each object's probability map. Similarly, the probability map for radar locations is generated by
\begin{equation*}
\resizebox{0.46\textwidth}{!}{$
    \mathcal{P}^r(\mathbf{x}) = \max_{j} \left\{ \mathcal{N} \left( \frac{1}{2 \pi \sqrt{|\mathbf{\Sigma}^r_{j}|}} \exp \left\{ -\frac{1}{2} (\mathbf{x} - \mathbf{\mu}^r_{j})^{\top} (\mathbf{\Sigma}^r_{j})^{-1} (\mathbf{x} - \mathbf{\mu}^r_{j}) \right\} \right) \right\},
$}
\end{equation*}
\begin{equation}
\resizebox{0.22\textwidth}{!}{$
    \mathbf{\mu}^r_{j} = 
    \begin{bmatrix}
    \rho^r_{j} \\
    \theta^r_{j} 
    \end{bmatrix}, 
    \mathbf{\Sigma}^r_{j} = 
    \begin{bmatrix}
    \delta^r_{j} & 0 \\
    0 & \epsilon(\theta^r_{j}) 
    \end{bmatrix}.
$}
\end{equation}
Here, $\delta^r_j$ is the radar's range resolution, and $\epsilon(\cdot)$ is the radar's azimuth resolution. 
Then, an element-wise product is used to obtain the fused probability map for each class,
\begin{equation}
    \mathcal{P}^{CRF}_{(cls)}(\mathbf{x}) = \mathcal{P}^c_{(cls)}(\mathbf{x}) * \mathcal{P}^r(\mathbf{x}).
\end{equation}
Finally, the fused annotations are derived from the fused probability maps $\mathcal{P}^{CRF}$ by peak detection.

\subsection{ConfMap Generation}
\label{subsec:confmap}

After objects are accurately localized in the radar range-azimuth coordinates, the annotations need to be transformed into a proper representation that is compatible with our RODNet. Considering the idea in \cite{cao2017realtime} that defines the human joint heatmap to represent joint locations, we define the confidence map (ConfMap) in range-azimuth coordinates to represent object locations. One set of ConfMaps has multiple channels, where each channel represents one specific class label. 
The value at the pixel in the $cls$-th channel represents the probability of an object with class $cls$ existing at that range-azimuth location. 
Thus, we use Gaussian distributions to set the ConfMap values around the object locations, whose mean is the object location, and the variance is related to the object class and scale information.

\section{Radar Object Detection}
\label{sec:rod}

\subsection{RODNet Architecture}
\label{subsec:rodnet}

\begin{figure} 
\centering
\includegraphics[width=.92\linewidth]{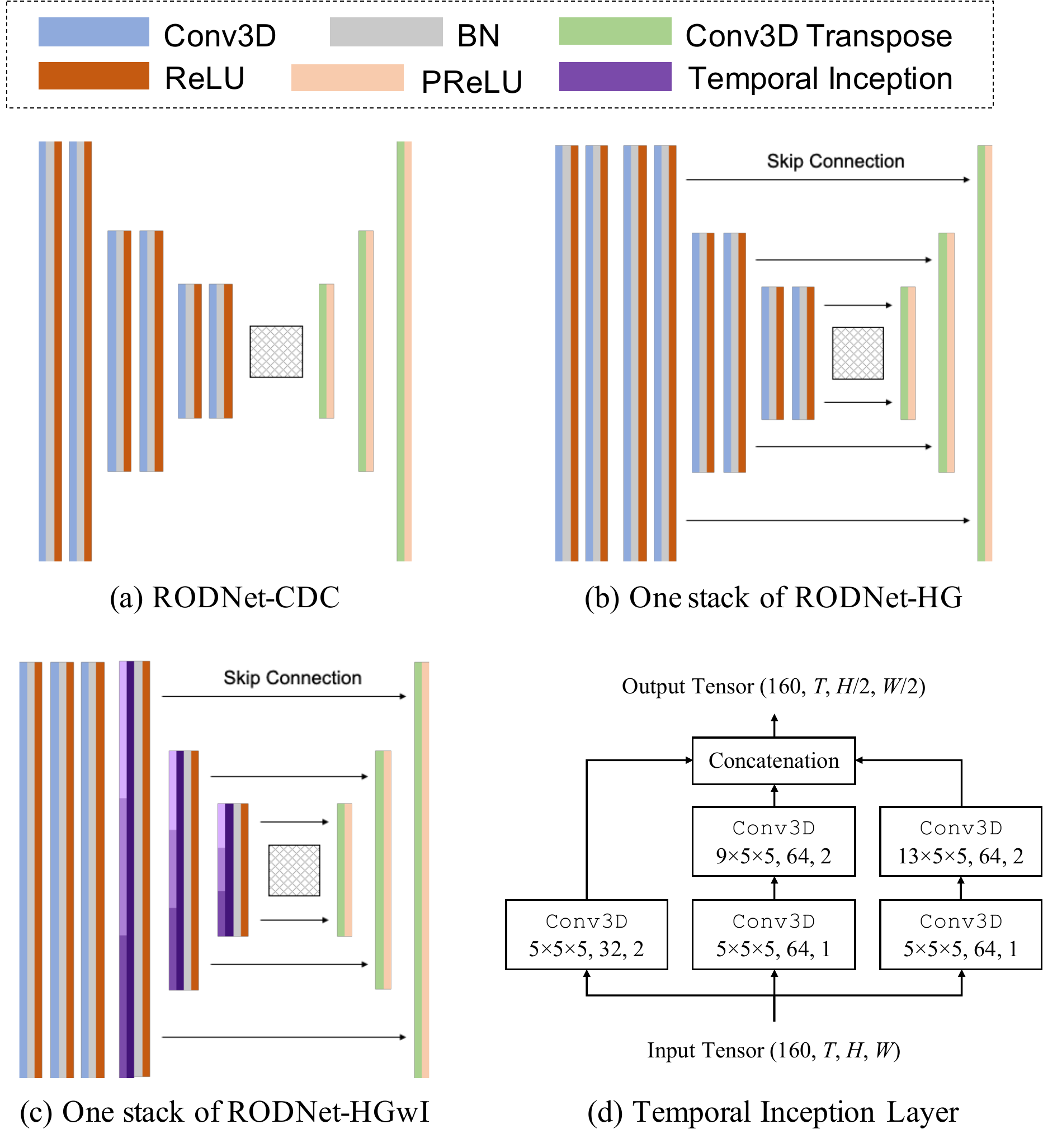}
  \caption{The architectures of our three RODNet models.}
  \label{fig:rodnet_arch}
\end{figure}

The three different network architectures for the RODNet are shown in Figure~\ref{fig:rodnet_arch}, named 3D Convolution-Deconvolution (RODNet-CDC), 3D stacked hourglass (RODNet-HG), and 3D stacked hourglass with temporal inception (RODNet-HGwI), respectively.
RODNet-CDC is a shallow 3D CNN network that squeeze the features in both spatial and temporal domains to better extract temporal information.  
While the RODNet-HG is adopted from \cite{newell2016stacked}, but we replace 2D convolution layers with 3D convolution layers and adjust the parameters for our task. 
As for the RODNet-HGwI, we replace the 3D convolution layers in each hourglass by the temporal inception layers \cite{szegedy2015going} with different temporal kernel scales $(5, 9, 13)$ to extract different lengths of temporal features from the RAMaps. 

Overall, our RODNet is fed with a snippet of RAMaps $\bm{R}$ with dimension $(C_{RF}, \tau, w, h)$ and predicts ConfMaps $\bm{\hat{D}}$ with dimension $(C_{cls}, \tau, w, h)$, where $C_{RF}$ is the number of channels in each RAMap, referring \cite{zhao2018through}, where real and imaginary values are treated as two different channels, i.e., $C_{RF} = 2$; $\tau$ represents the snippet length; $C_{cls}$ is the number of object classes; $w$ and $h$ are width and height of RAMaps or ConfMaps respectively. Thus, RODNet predicts separate ConfMaps for individual radar frames. With systematically derived CRF annotations, we train our RODNet using binary cross entropy loss,
\begin{equation}
    \ell = - \sum_{cls} \sum_{i,j} \bm{D}_{i,j}^{cls} \log \bm{\hat{D}}_{i,j}^{cls} + \left( 1 - \bm{D}_{i,j}^{cls} \right) \log \left( 1 - \bm{\hat{D}}_{i,j}^{cls} \right).
\end{equation}
Here, $\bm{D}$ represents the ConfMaps generated from camera annotations, $\bm{\hat{D}}$ represents the ConfMaps prediction, $(i,j)$ represents the pixel indices, and $cls$ is the class label.

\subsection{L-NMS: Identify Detections from ConfMaps}
\label{subsec:postprocess}

To obtain the final detections from ConfMaps, a post-processing step is still required. 
Here, we adopt the idea of non-maximum suppression (NMS), which is frequently used in image-based object detection to remove the redundant bounding boxes from the detectors. 
Here, NMS uses intersection over union (IoU) as the criterion to determine if a bounding box should be removed. However, there is no bounding box definition in our problem. Thus, inspired by object keypoint similarity (OKS) defined for human pose evaluation in COCO dataset \cite{lin2014microsoft}, we define a new metric, called object location similarity (OLS), to take the role of IoU, which describes the relationship between two detections considering their distance, classes and scale information on ConfMaps. More specifically, 
\begin{equation}
    \text{OLS} = \exp \left\{ \frac{-d^2}{2 (s \kappa_{cls})^2} \right\},
    \label{eq:ols}
\end{equation}
where $d$ is the distance (in meters) between the two points on RAMap, $s$ is the object distance from the sensors, representing object scale information, and $\kappa_{cls}$ is a per-class constant which represents the error tolerance for class $cls$, which can be determined by the object average size of the corresponding class. Moreover, we empirically tune $\kappa_c$ to make OLS distributed reasonably between 0 and 1. 
Here, we try to interpret OLS as a definition of Gaussian probability, where distance $d$ acts as bias and $(s \kappa_{cls})^2$ acts as variance. 
Therefore, OLS is a distance metric in a similarity manner, which also considers object sizes and distances, so that more reasonable than other traditional distance metrics, such as Euclidean distance, Mahalanobis distance, etc. 
This OLS metric is also used to \textit{match detections and ground truth} for evaluation purpose, mentioned in Section~\ref{subsec:evaluation}.

After OLS is defined, we propose a location-based NMS (L-NMS), whose procedure can be summarized as follows:
\begin{itemize}
\setlength{\itemsep}{1pt}
    \item[1)] Get all the peaks in all $C$ channels in ConfMaps within a $3 \times 3$ window as a peak set $P = \{p_n\}_{n=1}^{N}$.
    \item[2)] Pick the peak $p^* \in P$ with the highest confidence and remove it from the peak set. Calculate OLS with each of the rest peaks $p_i$, where $p_i \neq p^*$.
    \item[3)] If OLS between $p^*$ and $p_i$ is greater than a threshold, remove $p_i$ from the peak set.
    \item[4)] Repeat Steps 2 and 3 until the peak set becomes empty.
\end{itemize}


Moreover, during the inference stage, we can send overlapped RAMap snippets into the RODNet, which provides different ConfMaps predictions for a single radar frame. Then, we merge these different ConfMaps together to obtain the final ConfMaps results. 
This scheme can improve the system's robustness and can be considered as a performance-speed trade-off, discussed in Section~\ref{subsec:results}.

\section{CRUW Dataset}
\label{sec:dataset}

\begin{figure}[t]
\centering
\includegraphics[width=\linewidth]{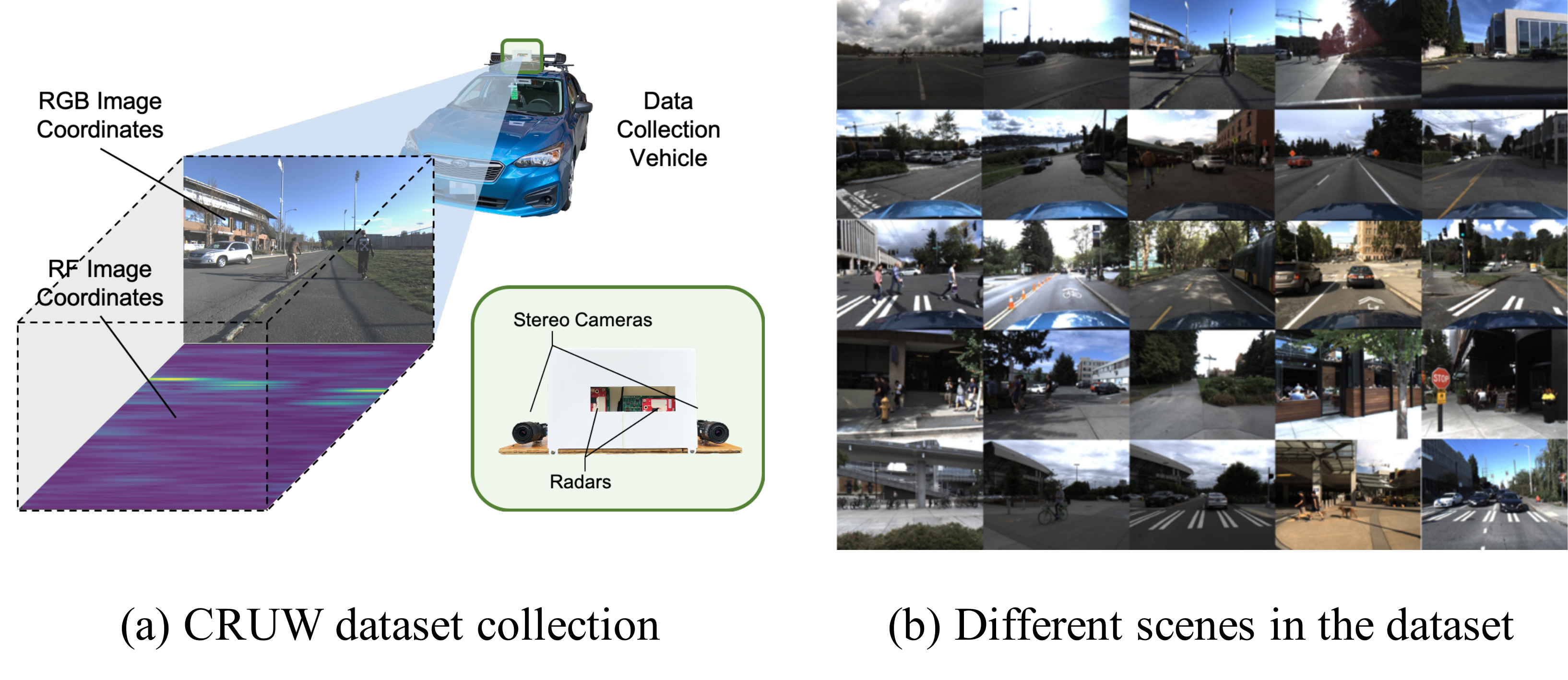}
  \caption{Sensor platform and driving scenes in CRUW dataset. 
  }
  \label{fig:dataset}
\end{figure}

\begin{figure}[t]
\centering
  \includegraphics[width=.94\linewidth]{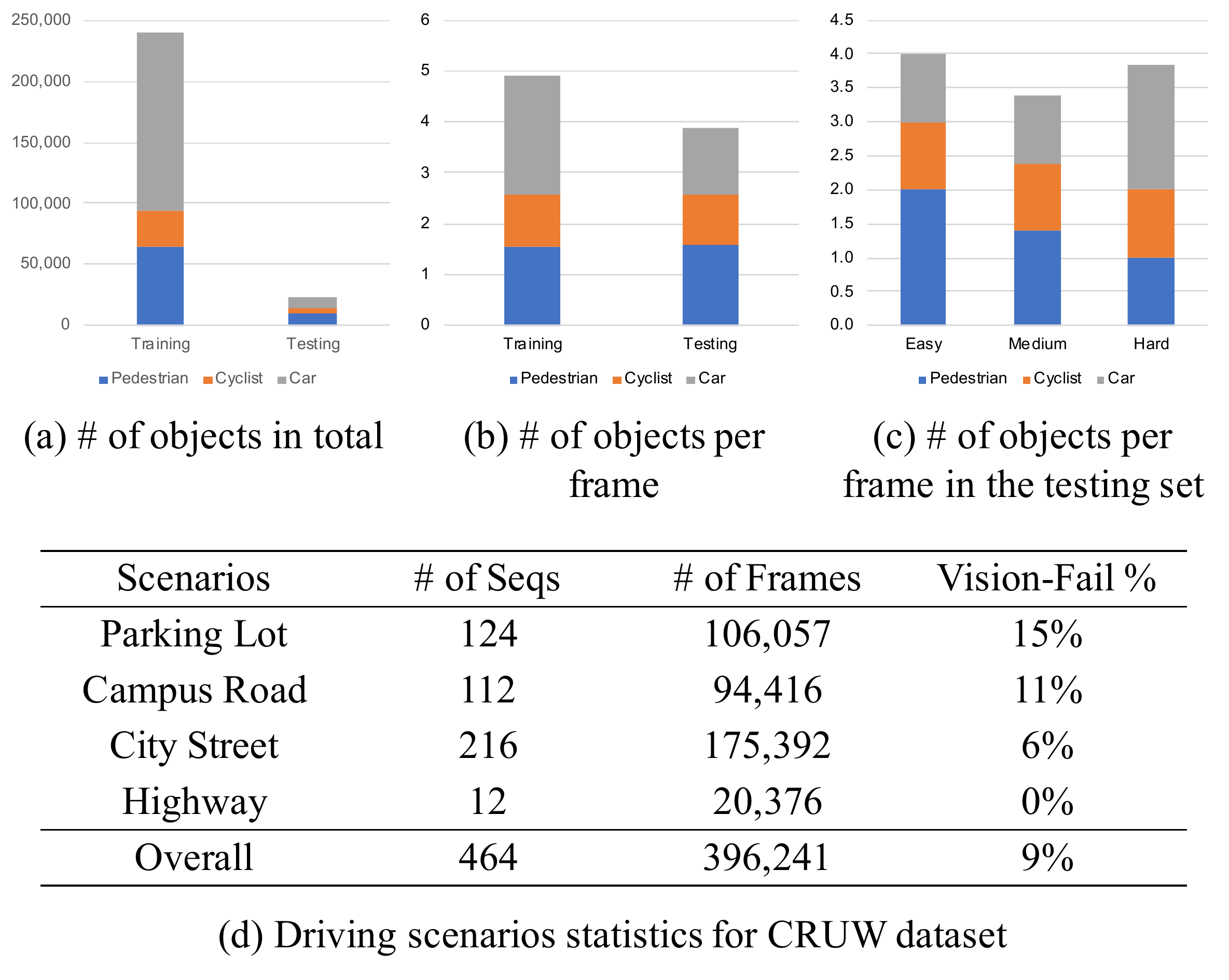}
   \caption{Illustration for our CRUW dataset distribution.}
\label{fig:cr_distribution}
\end{figure}

\begin{figure*}[t]
    \centering
    \includegraphics[width=.9\linewidth]{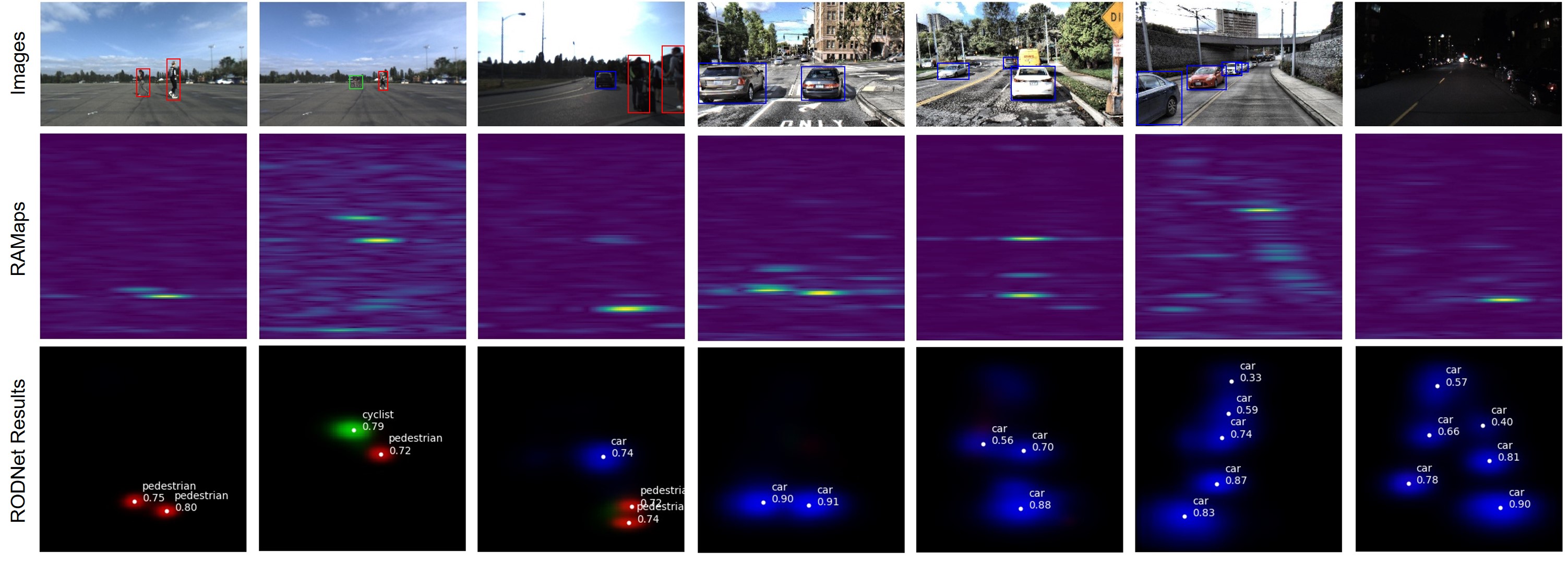}
    \captionof{figure}{Example results from our RODNet. The first row shows the images and the second row is the corresponding Radar frames in RAMap format. The ConfMaps predicted by the RODNet is shown in the third row, where the white dots represent the final detections after post-processing. Different colors in the ConfMaps represent different detected object classes.}
    \vspace{1em}
\label{fig:res_qual_vis}
\end{figure*}

Going through some existing autonomous driving datasets  \cite{geiger2013vision,apollo_scape_dataset,waymo_open_dataset,nuscenes2019,RadarRobotCarDatasetICRA2020}, only nuScenes \cite{nuscenes2019} and Oxford RobotCar \cite{RadarRobotCarDatasetICRA2020} consider radar. However, the format is 3D radar points, which are usually sparse and without motion and surface texture information that needed for our task.  
In order to efficiently train and evaluate our RODNet using RF data, we collect a new dataset -- CRUW dataset.
Our sensor platform contains a pair of stereo cameras \cite{flir} and two 77GHz FMCW radar antenna arrays \cite{ti}. The sensors, assembled and mounted together as shown in Figure~\ref{fig:dataset}~(a), are well-calibrated and synchronized. 
Even though our cross-modal supervision requires just one monocular camera, the stereo cameras are setup to provide some ground truth of depth for object 3D localization performance validation.

The CRUW dataset contains more than 3 hours with 30 FPS (about 400K frames) of camera/radar data under different driving scenarios, including campus road, city street, highway, parking lot, etc. Some sample visual data are shown in Figure~\ref{fig:dataset}~(b). 
Besides, we also collect several vision-fail scenarios where the image qualities are pretty bad, i.e., dark, strong light, blur, etc. These data are only used for testing to illustrate that our method can still be reliable when vision techniques fail. 

The object distribution in CRUW is shown in Figure~\ref{fig:cr_distribution}. The statistics only consider the objects within the radar field of view. There are about 260K objects in CRUW dataset in total, including $92\%$ training and $8\%$ testing. The average number of objects in each frame is similar between training and testing data. 
From each scenario, we randomly select several complete sequences as testing sequences, which are not used for training. Thus, the training and testing sequences are captured at different locations and different time to show the generalization capability of the proposed system.
For the ground truth needed only for evaluation purposes, we annotate $10\%$ of the visible data and $100\%$ vision-fail data. The annotation is operated on RAMaps by labeling the object classes and locations according to the corresponding images and RAMap reflection magnitude.

\section{Experiments}
\label{sec:experiments}

\subsection{Evaluation Metrics}
\label{subsec:evaluation}

To evaluate the performance, we utilize our proposed OLS (Eq.~\ref{eq:ols}), replacing the role of IoU in image-level object detection, to determine whether the detection result can be matched with a ground truth.
During the evaluation, we first calculate OLS between each detection result and ground truth in every frame. Then, we use different thresholds from $0.5$ to $0.9$ with a step of $0.05$, for OLS and calculate the average precision (AP) and average recall (AR) for all different OLS thresholds. Here, we use AP and AR to represent the average values among different OLS thresholds, and use $\text{AP}^{\text{OLS}}$ and $\text{AR}^{\text{OLS}}$ to represent the values at a certain OLS threshold. Overall, we use AP and AR as our main evaluation metrics for the radar object detection task.

\begin{table*}[t]
\footnotesize
\centering
\caption{Radar object detection performance evaluated on CRUW dataset. }
\begin{tabular}{L{3cm}|C{0.9cm}C{0.9cm}|C{0.9cm}C{0.9cm}|C{0.9cm}C{0.9cm}|C{0.9cm}C{0.9cm}}
\hline
\multirow{2}*{Methods} & \multicolumn{2}{c|}{Overall} & \multicolumn{2}{c|}{Easy} & \multicolumn{2}{c|}{Medium} & \multicolumn{2}{c}{Hard} \\
\cline{2-9}
~ & AP & AR & AP & AR & AP & AR & AP & AR  \\
\hline
Decision Tree \cite{gao2019experiments} & $4.70$ & $44.26$ & $6.21$ & $47.81$ & $4.63$ & $43.92$ & $3.21$ & $37.02$ \\
CFAR+ResNet \cite{8468324} & $40.49$ & $60.56$ & $78.92$ & $85.26$ & $11.00$ & $33.02$ & $6.84$ & $36.65$\\
CFAR+VGG-16 \cite{gao2019experiments} & $40.73$ & $72.88$ & $85.24$ & $88.97$ & $47.21$ & $62.09$ & $10.97$ & $45.03$ \\
\hline
\textbf{RODNet (Ours)} & $83.76$ & $85.62$ & $94.52$ & $95.94$ & $72.49$ & $75.59$ & $66.77$ & $71.24$ \\
\hline
\end{tabular}
\label{tab:results_main}
\end{table*}

\begin{table*}[t] 
\footnotesize
\centering
\caption{Ablation studies on the performance improvement with different architectures and annotations. }
\begin{tabular}{L{2.1cm}|C{1.6cm}||C{0.9cm}|C{0.9cm}C{0.9cm}C{0.9cm}||C{0.9cm}|C{0.9cm}C{0.9cm}C{0.9cm}}
\hline
Architectures & Supervision & AP & $\text{AP}^{0.5}$ & $\text{AP}^{0.7}$ & $\text{AP}^{0.9}$ & AR & $\text{AR}^{0.5}$ & $\text{AR}^{0.7}$ & $\text{AR}^{0.9}$\\
\hline
\multirow{2}{*}{RODNet-CDC} & CO & $52.62$ & $78.21$ & $54.66$ & $18.92$ & $63.95$ & $84.13$ & $68.76$ & $30.71$ \\
& CRF & $74.29$ & $78.42$ & $76.06$ & $64.58$ & $77.85$ & $80.05$ & $78.93$ & $71.72$ \\
\hline
\multirow{2}{*}{RODNet-HG} & CO & $73.86$ & $80.34$ & $74.94$ & $61.16$ & $79.87$ & $83.94$ & $80.73$ & $71.39$  \\
& CRF & $81.10$ & $84.71$ & $83.08$ & $70.21$ & $84.26$ & $86.54$ & $85.42$ & $77.44$ \\
\hline
\multirow{2}{*}{RODNet-HGwI} & CO & $77.75$ & $82.88$ & $79.93$ & $61.88$ & $81.11$ & $85.13$ & $82.78$ & $68.63$  \\
& CRF & $83.76$ & $87.99$ & $86.00$ & $70.88$ & $85.62$ & $88.79$ & $87.37$ & $76.26$\\
\hline
\end{tabular}
\label{tab:results_apolss}
\end{table*}

\subsection{Radar Object Detection Results}
\label{subsec:results}

We train our RODNet using the training data with CRF annotations in CRUW dataset. For testing, we perform inference and evaluation on the human-annotated visible data. 
The quantitative results are shown in Table~\ref{tab:results_main}. 
We compare our results with the following radar-only baselines: 1) a decision tree using handcrafted features from radar data \cite{gao2019experiments}; 2) a radar object classification network \cite{8468324} appended after the CFAR detection algorithm; 3) radar object detection method reported in \cite{gao2019experiments}. 
To evaluate the performance under different scenarios, we split the test set into three levels, i.e., easy, medium, and hard. 
Among all the three competing methods, the AR performance for \cite{gao2019experiments}, \cite{8468324} is relatively stable in the three different test sets, but their APs vary a lot. Especially, the APs drop from around $80\%$ to $10\%$ for easy to hard testing sets. This is caused by a large number of false positives detected by the traditional CFAR algorithm, which would significantly decrease the precision.
Comparing with the baseline and competing methods, our RODNet outperforms significantly on both AP and AR metrics, achieving the best performance of $83.76\%$ AP and $85.62\%$ AR with the RODNet-HGwI architecture and CRF supervision. From now on, the RODNet discussed is referring to RODNet-HGwI, unless specified.

Some qualitative results are shown in Figure~\ref{fig:res_qual_vis}, where we can see that the RODNet can accurately localize and classify multiple objects in different scenarios. The examples on the left of Figure~\ref{fig:res_qual_vis} are the scenarios that are relatively clean with fewer noises on the RAMaps, while the right ones are more complex with different kinds of obstacles, like trees, traffic sign, walls, etc. Especially, in the second to the last example, we can see high reflections on the right of the RAMap, which comes from the walls. The resulting ConfMap shows that the RODNet does not recognize them as any object, which is quite promising. 

\begin{figure}[t]
\centering
\includegraphics[width=0.68\linewidth]{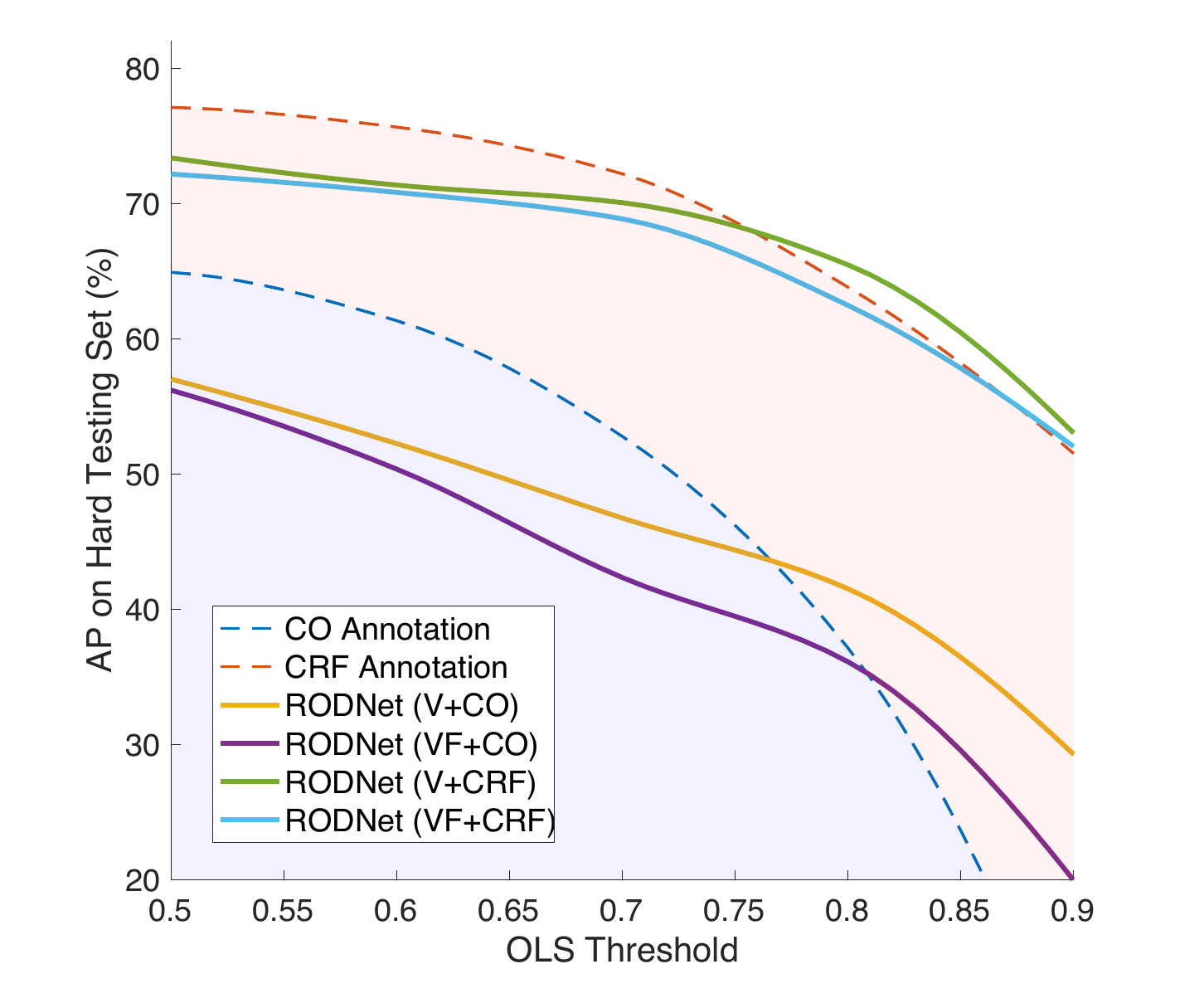}
  \caption{Performance of vision-based and our RODNet on ``Hard'' testing set with different localization error tolerance. (V: visible data; VF: vision-fail data)}
  \label{fig:results_vrcomp}
\end{figure}

\begin{table}[t]
    \footnotesize
    \centering
    \caption{The mean localization error (standard deviation) of CO/CRF annotations on CRUW dataset (in meters).}
    \begin{tabular}{c C{1.6cm} C{1.6cm} C{1.6cm}}
        \hline
        Supervision & Pedestrian & Cyclist & Car \\
        \hline
        CO & $0.69\ (\pm 0.77)$ & $0.87\ (\pm 0.89)$ & $1.57\ (\pm 1.12)$ \\
        CRF & $0.67\ (\pm 0.55)$ & $0.82\ (\pm 0.59)$ & $1.26\ (\pm 0.64)$ \\
        \hline
    \end{tabular}
    \label{tab:res_obj3dloc}
\end{table}

\subsection{Ablation Studies}
\label{subsec:ablation}


\begin{figure}[t]
\centering
\includegraphics[width=0.7\linewidth]{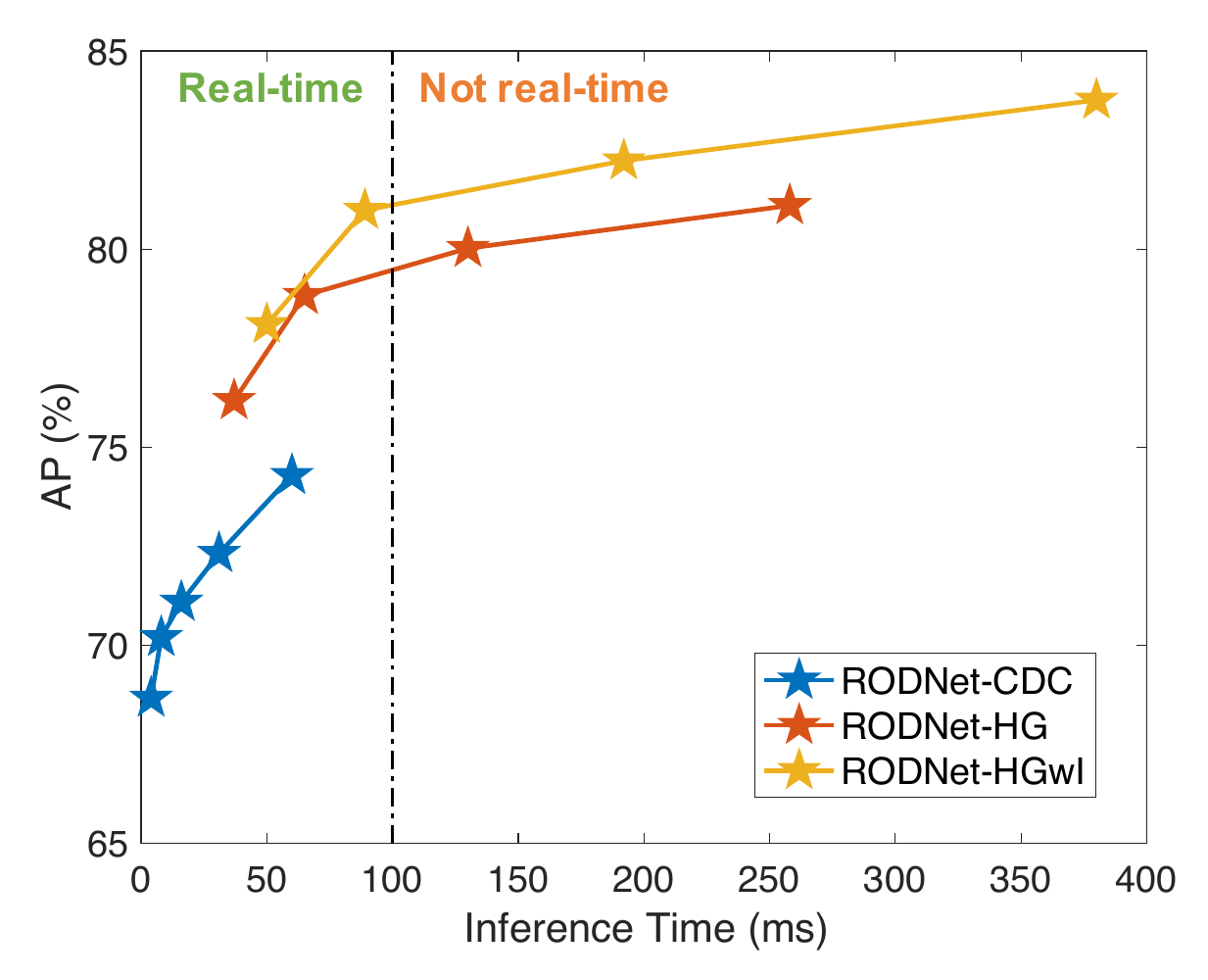}
  \caption{Performance-speed trade-off for the RODNet real-time implementation.}
  \label{fig:time_complx}
\end{figure}

First, AP and AR under different OLS thresholds are analyzed in Table~\ref{tab:results_apolss}. 
Besides, we compare the teacher's performance on object 3D localization for both CO and CRF annotations, shown in Table~\ref{tab:res_obj3dloc}. The CRF annotations are more accurate than CO annotations especially for the cars. From Table~\ref{tab:results_apolss} and \ref{tab:res_obj3dloc}, we can find that, with more robust CRF annotations, the performance of our RODNet can increase significantly for all the three architectures. 
In Figure~\ref{fig:results_vrcomp}, the performance of teacher and student are compared on ``Hard'' testing set. Our RODNet shows its superiority and robustness on its localization performance.

Second, real-time implementation is important for autonomous driving applications. As mentioned in Section~\ref{subsec:postprocess}, we use different overlapping lengths during the inference, running on an NVIDIA TITAN XP, and report the time consumed in Figure~\ref{fig:time_complx}. 
Here, we show the AP of three building architectures for the RODNet, and use 100~ms as a reasonable real-time threshold. 

\begin{figure*}[t] 
\centering
\includegraphics[width=0.8\linewidth]{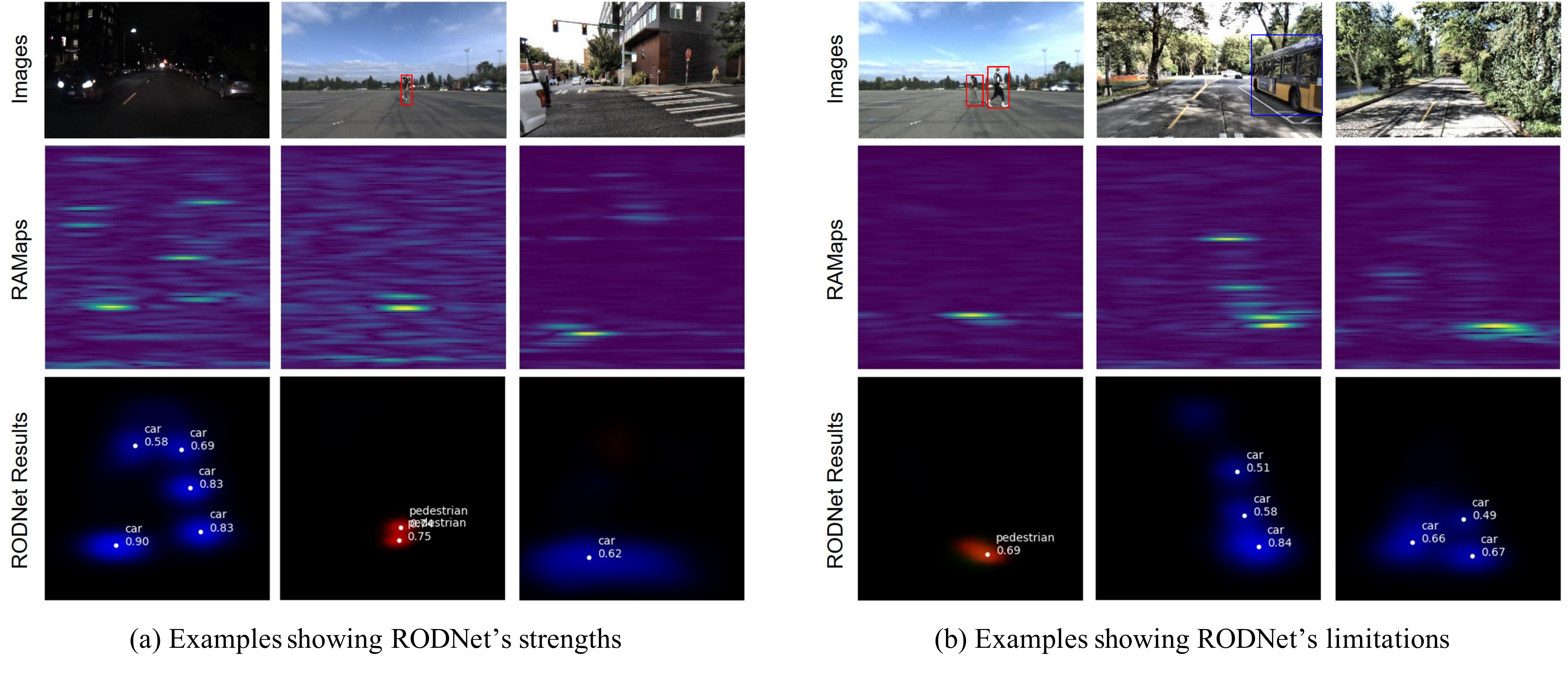}
  \caption{Examples illustrate strengths and limitations of our RODNet.}
  \label{fig:res_stren_limit}
\end{figure*}

\begin{wrapfigure}{r}{0.4\linewidth}
\centering
\includegraphics[width=\linewidth]{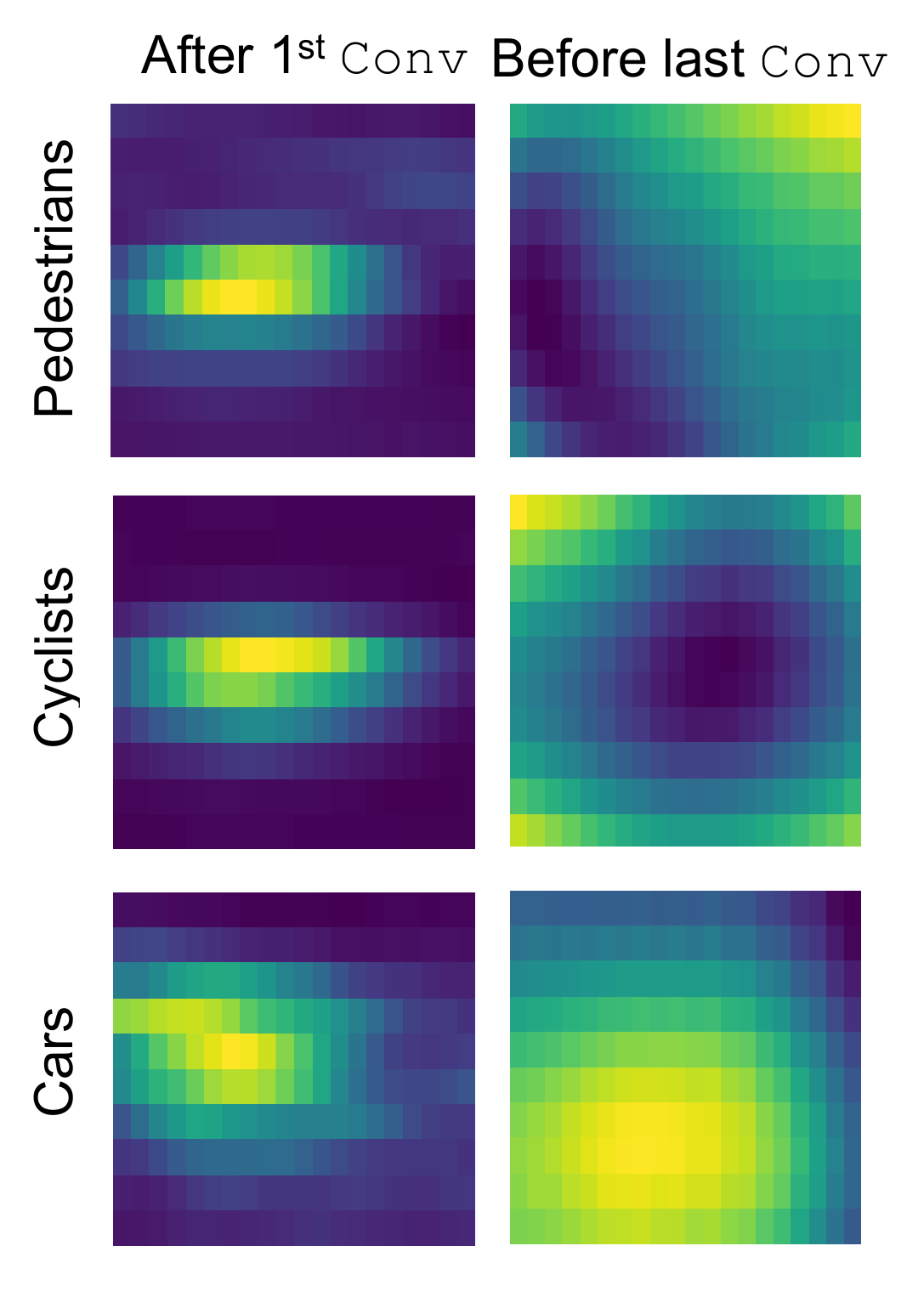}
  \caption{RODNet feature visualization. }
  \label{fig:feature_viz}
\end{wrapfigure}

After the RODNet is well-trained, we analyze the features learned from the radar data. In Figure~\ref{fig:feature_viz}, we show two different kinds of feature maps, i.e., the features after the first convolution layer and the features before the last layer. These feature maps are generated by cropping some randomly chosen objects from the original feature maps and average them into one. From the visualization, we notice that the feature maps are similar in the beginning, but they become more discriminative at the end of the RODNet. Note that the visualized features are pixel-wise averaged within each object class to better represent the general class-level features.

\subsection{Strengths and Limitations}
\label{subsec:stren_limit}

\paragraph{RODNet Strengths.} 
Some examples to illustrate the RODNet's advantages are shown in Figure~\ref{fig:res_stren_limit}~(a). First, the RODNet has similar performance in some severe conditions, like during the night, shown in the first example. Moreover, the RODNet can handle some occlusion cases when the camera usually fails. In the second example, two pedestrians are nearly fully occluded in the image, but our RODNet can still detect both of them. This is because they are separate in the radar point of view.
Last but not least, the RODNet has a wider field of view (FoV) than vision so that it can see more information. As shown in the third example, there is only a small part of the car visible in the camera view, which can hardly be detected from the camera side, but the RODNet can successfully detect it. 

\paragraph{RODNet Limitations.}
Some failure cases are shown in Figure~\ref{fig:res_stren_limit}~(b). When two objects are very near, the RODNet often fails to distinguish them due to the limited resolution of radar. In the first example, the RAMap patterns of the two pedestrians are intersected, so that our result only shows one pedestrian detected. Another problem is, for huge objects like bus and train, the RODNet often detects it as multiple objects as shown in the second example. Lastly, the RODNet is sometimes affected by noisy surroundings. In the third example, there is no object in the view, but the RODNet detects the obstacles as several cars. The last two problems should be solved with a larger training dataset.


\section{Conclusion}
\label{sec:conclusion}

Object detection is crucial in autonomous driving and many other areas. Computer vision society has been focusing on this topic for decades and come up with many good solutions. However, vision-based detection is still suffering from many severe conditions. This paper proposed a brand-new and novel object detection method purely from radar information, which is more robust than vision. The proposed RODNet can accurately and robustly detect objects in various autonomous driving scenarios even during the night or bad weather. Moreover, this paper presented a new way to learn radar data using cross-modal supervision, which can potentially improve the role of radar in autonomous driving applications.

\section*{Acknowledgement}
This work was partially supported by CMMB Vision -- UWECE Center on Satellite Multimedia and Connected Vehicles. The authors would also like to thank the colleagues and students in Information Processing Lab at UWECE for their help and assistance on the dataset collection, processing, and annotation works.

{\balance
{\small
\bibliographystyle{ieee_fullname}
\bibliography{egbib}
}}

\end{document}